\pdfoutput=1
\documentclass[11pt]{article}

\usepackage{acl}
\usepackage{algorithm}
\usepackage{multirow}
\usepackage{times}
\usepackage{latexsym}
\usepackage{graphicx}
\usepackage{booktabs}

\usepackage[T1]{fontenc}
\usepackage[utf8]{inputenc}

\usepackage{microtype}

\title{Logographic Information Aids Learning Better Representations\\ for Natural Language Inference}

\author{Zijian Jin \\
Tandon School of Engineering\\
  New York University \\
  \texttt{zj2076@nyu.edu} \\\And
  Duygu Ataman \\
  Courant Institute of Mathematical Sciences\\
  New York University \\
  \texttt{ataman@nyu.edu} \\}

\begin{document}
\maketitle
\begin{abstract}
Statistical language models conventionally implement representation learning based on the contextual distribution of words or other formal units, whereas any information related to the logographic features of written text are often ignored, assuming they should be retrieved relying on the cooccurence statistics. 
On the other hand, as language models become larger and require more data to learn reliable representations, such assumptions may start to fall back, especially under conditions of data sparsity. Many languages, including Chinese and Vietnamese, use logographic writing systems where surface forms are represented as a visual organization of smaller graphemic units, which often contain many semantic cues. In this paper, we present a novel study which explores the benefits of providing language models with logographic information in learning better semantic representations. We test our hypothesis in the natural language inference (NLI) task by evaluating the benefit of computing multi-modal representations that combine contextual information with glyph information. Our evaluation results in six languages with different typology and writing systems suggest significant benefits of using multi-modal embeddings in languages with logograhic systems, especially for words with less occurence statistics.%, suggesting multi-modality in representation learning is a promising direction to address data sparsity.
\end{abstract}

\section{Introduction}

The essential idea in statistical language modeling is to represent the meaning of a word as a function of its context. The function, modeled via the conditional probability of observing a word in a given utterance, has most efficiently been approximated with a neural network based architecture \citep{DBLP:journals/corr/abs-1301-3781,mikolov2013distributed,bengio2003neural,rnnlm,sundermeyer2012lstm}. The outstanding performance of neural methods in language modeling and their recent development \citep{peters2018deep,tenney2019bert} have them a preliminary component in various downstream NLP tasks. 

One of the main limitations in the formulation of language models lies however in the choice of ortographic units in calculating the contextual distribution, which is usually convenient in English and other languages using phonetic scripts. On the other hand, many languages rely on logographic writing systems, where surface forms are represented as a visual organization of smaller graphemic units and the word meaning can be changed through compositional variations of these units. Although a direct segmentation of these units has been found quite challenging due to visual compositions in the final form of the grapheme, previous studies have found potential benefits of using visual information to aid NLP models in sentence representation \cite{liu2017learning,meng2019glyce,DBLP:conf/emnlp/DaiC17,salesky2021robust}. On the other hand, none of these studies have focused on isolating the effects of different linguistic features in relation to their correlation to visual features.

\begin{figure}[t]
    \centering
    %\vspace{-1mm}
    \scalebox{0.13}{
    \includegraphics{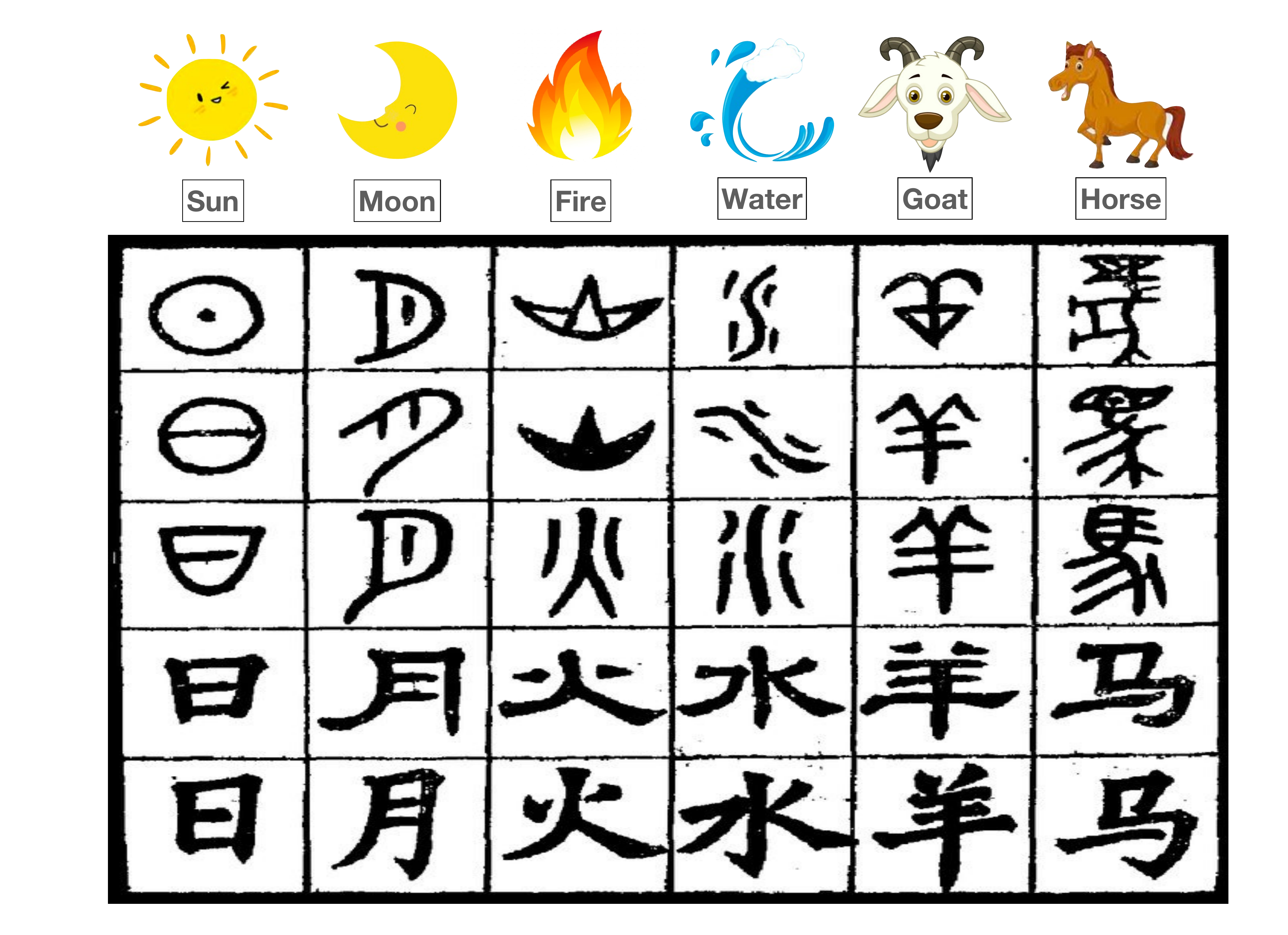}}
    \vspace{-3mm}
    \caption{Logographic information in Chinese.}
    \label{fig:Logoeg}
    %\vspace{-4mm}
\end{figure}

As shown in Figure~\ref{fig:Logoeg}, logographic information often contain important features related to the word meaning. In this paper, we perform the first focused analysis to measure the significance of logographic features specifically to the semantic information encoded in token or character-level language representations by evaluating the performance of multi-modal embeddings in the NLI task. In particular, we aim to answer the following research questions:
\begin{itemize}
  \item [1)] 
  How important may logographic information be to for an accurate representation of semantic information in word or character-level language units 
  \item [2)]
 Whether the contribution of logographic information to semantic representations may depend on the language typology and writing system
\end{itemize}

In order to answer these questions we implement a multi-modal representation learning model where each written text segment is representation as a combination of visual embeddings obtained from prominent convolutional neural network (CNN) based models \cite{liu2017learning,meng2019glyce,salesky2021robust}, and contextual representations obtained from multilingual pre-trained language models \cite{DBLP:conf/naacl/DevlinCLT19,DBLP:conf/acl/ConneauKGCWGGOZ20}. We evaluate the contribution of visual information to the performance in the NLI task under few-shot learning settings in six languages with varying typology and writing systems: English, Spanish, Hindi, Urdu, Vietnamese and Chinese. We also study the optimal representation granularity for semantic information by comparing word or character-level multi-modal representations in our experiments.

\begin{figure*}[!htpb]
    \centering
    \vspace{-10mm}
    \scalebox{0.45}{
    \includegraphics{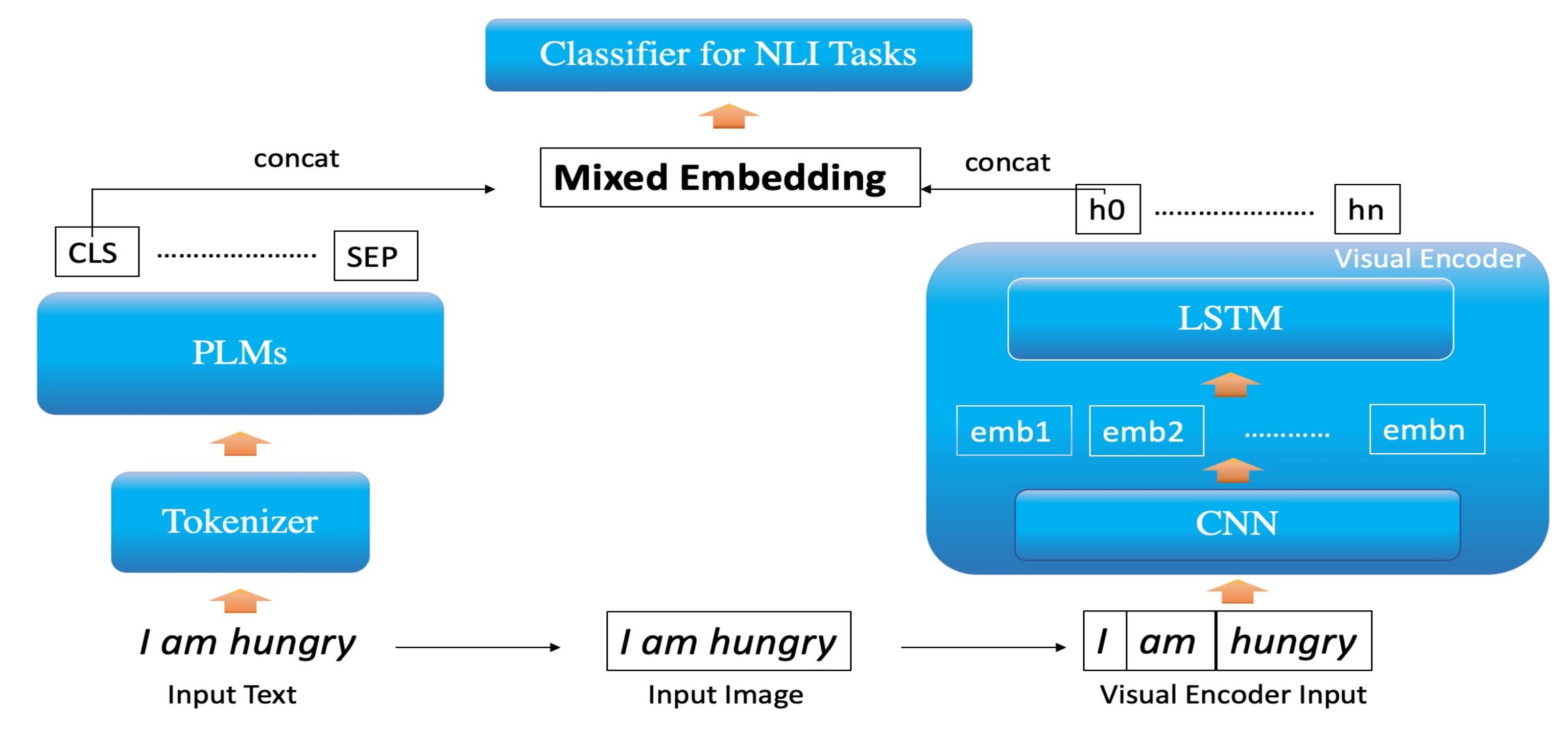}}
    \vspace{-3mm}
    \caption{Method overview.}
    \label{fig:pipeline}
    \vspace{-4mm}
\end{figure*}

In conclusion, we find that taking into account the visual information improves the performance in NLI tasks especially in logographic languages like Chinese and that the improvements are correlated with the factors that determine the quality of token representations, such as the occurence of the tokens in training data as well as language model capacity and hyperparameters. Our findings suggest multi-modal processing is a promising direction, especially for processing languages where conditions of data sparsity may create fall backs in assumptions undertaken in statistical formulations. 
%outperforms than  the previous state of the art on a series of tasks by introducing new method for learning cross-lingual representations. With training RoBERTa on a huge, multilingual dataset at an enormous scale, XLM-R \citep{DBLP:conf/acl/ConneauKGCWGGOZ20} leads to significant performance gains for a wide range of cross-lingual transfer tasks. In this paper we use mBERT and XLM-R as our baselines.

\section{Computing Visual Glyph Embeddings}

Our multi-modal embedding model is composed of two components: \textit{(i) the visual encoder}, which computes embeddings based on the input images representing each text segment, and \textit{(ii) the pre-trained language model} providing the text-based embeddings.

%Our overall model composed with three different part, Images Converter, Visual Embedder and Combiner. 
\paragraph{Image conversion}
%突出有两种区分方法？
%In this part we first turn the 
Text segments consisting of complete sentences are split into words (or characters) and then converted into images. Sentences are split into 30 x 60 pixel word images using the Jieba\footnote{https://github.com/fxsjy/jieba} tool. All graphemes are centralized to the middle of the image.

\paragraph{Visual Embeddings}
In order to extract the glyph information from text images, we use the CNN model developed by \citep{DBLP:conf/acl/LiuLLN17, DBLP:conf/nips/SutskeverVL14}  to generate visual embeddings. The model consists of a three-layer CNN, augmented with a two-layer feed-forward network. The full details of the network is given below.

\begin{table}
\begin{tabular}{c|c}
\hline \textbf{ Layer } & \textbf{Visual encoding model} \\
\hline 1 & \texttt { Spatial Conv. }(3,3) $\rightarrow$ 32 \\
2 & \texttt { ReLu } \\
3 & \texttt { MaxPool }(2,2) \\
4 & \texttt { Spatial Conv. }(3,3) $\rightarrow$ 32 \\
5 & \texttt { ReLu } \\
6 & \texttt { MaxPool }(2,2) \\
7 & \texttt { Spatial Conv. }(3,3) $\rightarrow$ 32 \\
8 & \texttt { ReLu } \\
9 & \texttt { Linear }(800,128) \\
10 & \texttt { ReLu } \\
11 & \texttt { Linear }(128,128) \\
12 & \texttt { ReLu } \\
\hline
\end{tabular}

\end{table}
%\label{fig-cnn}

The visual features extracted by the CNN are further encoded in a long-short term memory (LSTM) \cite{hochreiter1997long} network to learn the glyph embeddings. 

For a sequence consisting of $t$ tokens $x_0, x_1, ... , x_t$, the \textit{visual embedding} $v$ is computed by concatenating \citep{DBLP:conf/iclr/SuZCLLWD20, DBLP:conf/nips/LuBPL19} the hidden states of the LSTM and averaging them as

\[ 
v = \textrm{mean}([h_0; h_1; ...; h_t])
\]

\paragraph{Embedding composition}
In order to isolate the learning of representations from two modalities and measure their effect on the learning task in a controlled setting, we deploy late fusion in combining the visual embeddings with the text embeddings obtained by the pre-trained model for prediction in the down-stream task. The two embeddings are linearly composed through a simple affine projection and then concatenated. 
For the down-stream prediction task we use a multi-layer perceptron classifier.

\section{Experiments}

\subsection{Character recognition}

As an initial verification, we implement the visual encoder and evaluate it individually in the character recognition task. We use the CASIA Chinese Handwriting Database \cite{liu2011casia} and obtain competitive results (93.23\% accuracy) on this task, confirming the visual encoder works sufficiently in extracting character features from input images.

\subsection{NLI}

\paragraph{Data}

We evaluate our model under few-shot learning settings using the XNLI dataset \cite{conneau2020xnli}, using only a small portion of the testing data for training and development, and test the effect of logographic information to contribute to resolve the high level of semantic ambiguity. 
\begin{table}[ht]
	%\vspace{-2mm}
% 	\resizebox{1\columnwidth}{!}{
		\begin{tabular}{l|c}
			\hline
			\textbf{Datasets} & \textbf{Number of Sentences} \\
			\hline
			\textbf{Training} & 4509 \\
			\textbf{Development} & 501 \\
			\textbf{Test} & 2490 \\
			\hline
		\end{tabular}
		\caption{Data statistics for training, development and test sets.}
	\label{tab:task_statistics}
		%\vspace{-2.mm}
\end{table}

\paragraph{Model settings and hyper-parameters}
In training the multi-modal models, the learning rates of both XLM-R and mBERT based pre-trained models are set to 1e-6. The visual encoder is trained on the images captured from the training sentences, either at word or character-level resolution, with a learning rate of 4e-6 (for XLM-Roberta) and 1e-6 (for mBERT).
The hidden size of the LSTMs used is 128 and we use dropout of with 0.3 in this layer. All hyper-parameters are tuned with grid-search. For each task we train 30 epochs and always choose the results with smallest validation loss.

\paragraph{Languages} We pick six languages with varying typology and writing systems, including English, Spanish, Urdu, Vietnamese, Chinese and Hindi. \textbf{English} and \textbf{Spanish} use the Latin script; \textbf{Urdu} is written with the Arabic alphabet, whereas \textbf{Hindi} uses Devanagari, all of which are phonetic writing systems. \textbf{Chinese} uses logographic writing. \textbf{Vietnamese}, although traditionally have used logographic writing, recently and in the XNLI data set is written with the Latin script.

\paragraph{Contextual representations}
We verify the significance of logographic information for contributing to enrich the language representations by testing our multi-modal approach with two different pre-trained language models, including the \texttt{mBert-base} and the \texttt{XLM-R-base} both available from Huggingface\footnote{https://huggingface.co}. We also investigate the effects of different segmentation methods for processing sentence images either at the level of words or characters.

\section{Results and Discussion}

\begin{table*}[ht]
	\caption{Results in the XNLI benchmark. \texttt{base} models represent baseline pre-trained language model performance in the down-stream task. \texttt{base-CNN} models represent the multi-modal system performance. \texttt{(C)} denotes character and \texttt{(W)} denotes word level input representations. \texttt{Random} stands for comparisons to multi-modal systems where random images were input to the visual encoder to verify the effect of visual information on the overall performance.}
	\label{tab:task_results}
	\vspace{-2mm}
	\resizebox{2\columnwidth}{!}{
		\begin{tabular}{c|c|c|c|c|c|c}
			\hline
			\toprule
			\textbf{Languages} & \textbf{English} & \textbf{Chinese} & \textbf{Urdu} & \textbf{Hindi} & \textbf{Vietnamese} & \textbf{Spanish}\\
			\hline
			\textbf{mBERT-base} & \textbf{65.86} & 55.28 & 51.29 & 56.48 & 57.08 & \textbf{62.27}\\
			\textbf{mBERT-base-CNN (W)} & 62.87 & 58.88 & 53.49 & \textbf{57.68} & 59.88 & 60.47 \\
			\textbf{mBERT-base-CNN (C)} & 64.07 & \textbf{59.08} & \textbf{53.69} & 57.48 & \textbf{60.07} & 61.67 \\
			\textbf{mBERT-base-CNN (C) — Random} & - & 54.33 & - & - & - & -\\

			\textbf{XLM-Roberta-base} & \textbf{69.86} & 64.27 & \textbf{59.88} & \textbf{63.87} & 63.07 & \textbf{65.66}\\
			\textbf{XLM-Roberta-base-CNN (W)} & 69.26 & \textbf{66.66} & 57.88 & 62.87 & 61.67 & 65.46 \\
			\textbf{XLM-Roberta-base-CNN (C)} & 68.26 & 62.07 & 56.28 & 61.67 & \textbf{63.07} & 65.26 \\
			\textbf{XLM-Roberta-base-CNN (C) — Random} & - & 61.36 & - & - & - & -\\
			\bottomrule
			\hline
		\end{tabular}}
		%\vspace{-2.mm}
\end{table*}

Our experiment results are given in Table \ref{tab:task_results}. At a first glance, we observe the performance of the models are much lower than reported in \cite{conneau2020xnli}, since we have significantly less training and development data available. Under these challenging evaluation settings with high amount of sparsity, we observe that the logographic information improves the performance obtained using the \texttt{mBert-base} model in all languages that do not deploy the Latin script, including Chinese, Urdu, Hindi and Vietnamese. In case of \texttt{XLM-Roberta-base}, which had better optimization on a larger corpus, the overall performance are consistently better than \texttt{mBert-base} and the improvements remain consistent, especially in Chinese and Vietnamese. We hypothesize that the slightly higher amount of improvements in \texttt{mBERT-base} might be due to better quality of representations provided with the optimized training regime of \texttt{XLM-Roberta-base}.
Using the \texttt{mBERT-base} model, we find more advantage of embedding logographic information at the character level in Chinese, Urdu and Hindi, however, in Hindi, the results are comparable. When using the \texttt{XLM-Roberta-base}, we observe improvements in Chinese with word-level glyph embeddings and in Vietnamese using character-level glyph embeddings. While character-level embeddings might be suitable for a phonetic language like Vietnamese, the logograhic writing system in Chinese might make word-level visual embeddings more convenient, since the intra-graphemic dependencies can be captured at the visual level.

%The overall results suggest that logographic information can be useful in a variety of settings, including different languages. Naturally, the saturation level of parameters in the pre-trained model are essential in determining the final amount of fine-tuning possible, thus, this would determine the overall performance gains, although we find consistent improvements even in highly optimized models like the \texttt{XLM-Roberta}.

Although the improvements highly correlate with the logograhic nature of the writing system, the fact that they apply to most languages, even Urdu and Hindi with phonetic alphabets, point to the suboptimal effects in tokenization or segmentation and their potential harms to correctly model the contextual distribution. We also see in high-resource language representations like English, our fusion method may be harmful to the downstream task, which we anticipate that could be resolved with higher amount of fine-tuning and development data. In light of all these considerations, the findings suggest that multi-modality is a promising direction for overcoming problems related to data sparsity, and eventually tokenization or segmentation-free language modeling.

\begin{table}[ht]
	%\vspace{-2mm}
% 	\resizebox{1\columnwidth}{!}{
		\begin{tabular}{l|c|c}
			\hline
			\textbf{Models} & \textbf{Accuracy} & \textbf{\# of UNK} \\
			\hline
			\textbf{mBERT-base (C)} & 45.31 & \multirow{2}{*}{128} \\
			\textbf{mBERT-base-CNN (C)} & 52.43 & ~ \\
			\hline
		\end{tabular}
		\caption{Results for targeted evaluation, UNK represents unknown tokens.}
	\label{tab:targeted_result}
		%\vspace{-2.mm}
\end{table}
As an additional analysis investigating the effects of token frequency on the positive effects of logographic information integrated in the language model, we sample sentences in the test set that have unknown words in the model vocabulary and compute the targeted accuracy on this sample of sentences. The results shows in table~\ref{tab:targeted_result} further illustrate the boosted performance on the sample test, suggesting that data sparsity is an important obstacle to learning high-quality contextual representations, and such conditions can be the ideal place where logographic information might be useful to improve the semantic features embedded in representations.

\section{Conclusion}

In this paper, we evaluated the benefits of using logographic information in language modeling by implementing a multi-modal representation learning model which combines contextual language representations with visual embeddings. Our experiments in the NLI task in six languages confirmed the benefits of logograhic information in obtaining more reliable semantic representations, especially under sparse learning settings. As future work we hope to contribute to the development of larger multilingual benchmarks to evaluate the effect of visual information on more languages and linguistic phenomena. Our software and the experimental data will be available upon publication.

%\subsection{References}

%\subsection{Appendices}

%Use \verb|\appendix| before any appendix section to switch the section numbering over to letters. See Appendix~\ref{sec:appendix} for an example.

\section*{Acknowledgements}

The authors would like to express deep gratitude to Chinmay Hedge and Aishwarya Kamath for discussions and their useful feedback.
This work was supported by Samsung Advanced Institute of Technology (under the project Next Generation Deep Learning: From Pattern Recognition to AI) and NSF Award 1922658 NRT-HDR: FUTURE Foundations, Translation, and Responsibility for Data Science.
\bibliography{anthology,custom}

%\appendix

%\section{Example Appendix}
%\label{sec:appendix}

%This is an appendix.

\end{document}